Original Paper

# Cancer Diagnosis Categorization in Electronic Health Records Using Large Language Models and BioBERT: Model Performance Evaluation Study

Soheil Hashtarkhani[1], PhD; Rezaur Rashid[1], PhD; Christopher L Brett[2], MD; Lokesh Chinthala[1], MSc; Fekede Asefa Kumsa[1], PhD; Janet A Zink[1], PhD; Robert L Davis[1], MPH, MD; David L Schwartz[1,3], MD; Arash Shaban-Nejad[1], MPH, PhD

[1]Center for Biomedical Informatics, Department of Pediatrics, College of Medicine, University of Tennessee Health Science Center, Memphis, TN, United States
[2]University of Tennessee Graduate School of Medicine, Knoxville, TN, United States
[3]Departments of Radiation Oncology and Preventive Medicine, College of Medicine, University of Tennessee Health Science Center, Memphis, TN, United States

**Corresponding Author:**
Arash Shaban-Nejad, MPH, PhD
Center for Biomedical Informatics, Department of Pediatrics
College of Medicine, University of Tennessee Health Science Center
50 N Dunlap Street
Memphis, TN 38103
United States
Phone: 1 9012875836
Email: ashabann@uthsc.edu

## Abstract

**Background:** Electronic health records contain inconsistently structured or free-text data, requiring efficient preprocessing to enable predictive health care models. Although artificial intelligence–driven natural language processing tools show promise for automating diagnosis classification, their comparative performance and clinical reliability require systematic evaluation.

**Objective:** The aim of this study is to evaluate the performance of 4 large language models (GPT-3.5, GPT-4o, Llama 3.2, and Gemini 1.5) and BioBERT in classifying cancer diagnoses from structured and unstructured electronic health records data.

**Methods:** We analyzed 762 unique diagnoses (326 International Classification of Diseases [ICD] code descriptions, 436 free-text entries) from 3456 records of patients with cancer. Models were tested on their ability to categorize diagnoses into 14 predefined categories. Two oncology experts validated classifications.

**Results:** BioBERT achieved the highest weighted macro $F_1$-score for ICD codes (84.2) and matched GPT-4o in ICD code accuracy (90.8). For free-text diagnoses, GPT-4o outperformed BioBERT in weighted macro $F_1$-score (71.8 vs 61.5) and achieved slightly higher accuracy (81.9 vs 81.6). GPT-3.5, Gemini, and Llama showed lower overall performance on both formats. Common misclassification patterns included confusion between metastasis and central nervous system tumors, as well as errors involving ambiguous or overlapping clinical terminology.

**Conclusions:** Although current performance levels appear sufficient for administrative and research use, reliable clinical applications will require standardized documentation practices alongside robust human oversight for high-stakes decision-making.

# Introduction

Electronic health records (EHRs) generate vast volumes of data, providing a valuable resource for clinical decision-making and policy development [1]. Today, many prediction and prognostic models embedded within EHRs aid clinicians and policymakers by delivering data-driven insights [2,3]. However, a significant portion of the data in EHRs is either categorized inconsistently or stored as free text, requiring extensive manipulation before it can be utilized effectively in predictive models to support decision-making. This manual structuring process is often labor-intensive, time-consuming, and error-prone, especially when dealing with large datasets or ambiguous language [4]. Automating data structuring through natural language processing (NLP) is, therefore, essential to scale predictive health care applications.

With the advent of generative artificial intelligence (AI) and large language models (LLMs) [5], there has been a transformative leap in the capacity to process and interpret text with near-human accuracy. LLMs are advanced AI systems that use vast datasets and sophisticated transformer-based neural networks to understand and generate human language [6]. By leveraging deep learning, these models capture nuances in syntax, semantics, and context, supporting a range of language-related tasks such as text completion, translation, and summarization. This has positioned LLMs as powerful tools for automating the structuring of unstructured data [7]. Through application programming interface access, LLM capabilities can be applied to large datasets, including in real-time applications. However, ensuring the accuracy and reliability of these models is critical [8].

Although there have been several efforts to apply NLP to EHR data and cancer diagnosis [9-11], few studies have directly compared emerging general-purpose LLMs with specialized models like BioBERT (where BERT stands for bidirectional encoder representations from transformers). In particular, there is limited research evaluating their performance in a clinically relevant setting that includes both structured International Classification of Diseases (ICD) codes and unstructured free-text diagnoses. This study addresses that gap by benchmarking multiple LLMs and BioBERT using real-world oncology data, with expert validation to assess clinical relevance.

In this context, this study examines the potential of 4 LLMs—GPT-3.5, ChatGPT-4o, Llama 3.2, and Gemini 1.5—along with BioBERT, a specialized biomedical language model, to structure extracted EHR data in a case study focused on radiation therapy. Although the LLMs are general-purpose models capable of diverse language tasks, BioBERT is a domain-specific model pretrained on biomedical texts. We evaluate the effectiveness of these models in classifying patient diagnoses in ICD and free-text formats, with expert oncology review to validate model outputs. Our primary aim is to assess their capabilities for EHR data structuring, which is crucial for incorporating unstructured medical data into predictive models that forecast interruptions in radiation therapy.

# Methods

## Dataset

This study uses a dataset comprised of records of 3456 patients with cancer from the Research Enterprise Data Warehouse, originally sourced from the University of Tennessee Medical Center in Knoxville from 2017 to 2021. This dataset forms part of a larger research project focused on predicting interruptions in radiation therapy. It includes detailed demographic data (age, gender, race), administrative information (treatment dates, insurance type), and social determinants of health (eg, socioeconomic indicators), which are critical for understanding factors influencing treatment continuity.

Among these patients, 720 lacked standardized ICD codes for their cancer diagnoses, leaving only unstructured text descriptions from treatment plan forms. This gap offered a unique opportunity to evaluate the effectiveness of LLMs in structuring unstructured data. Examples of structured ICD codes and descriptions and unstructured free-text diagnoses include the following:

- ICD descriptions: "Malignant neoplasm of lower lobe, right bronchus or lung," "Benign neoplasm of cerebral meninges," "Malignant neoplasm of glottis."
- Free-text descriptions: "56-year-old female with an anal canal/perianal squamous cell carcinoma," "66-year-old male with metastatic esophageal adenocarcinoma," "b/l GBM," "left RMT."

## Category Definition

To ensure clinically relevant cancer type categorizations, a comprehensive literature review was conducted, followed by a focus group session with two oncology experts (DLS and CLB). This collaborative process resulted in the identification of 14 clinically relevant categories for this study: "Benign," "Breast," "Lung or Thoracic," "Prostate," "Gynecologic," "Head and Neck," "Gastrointestinal," "Central Nervous System," "Metastasis," "Skin," "Soft Tissue," "Hematologic," "Genitourinary," and "Unknown." These categories were designed to align with the clinical requirements of oncology treatment planning and the predictive modeling of radiation therapy interruptions.

Unlike standard ICD ontologies, this categorization was customized for the specific needs of the project. For example, all benign tumors were grouped into a single category, regardless of location or ICD hierarchy, to reflect their distinct impact on treatment interruption. Therefore, we required language models capable of interpreting both ICD code descriptions and free-text entries and classifying them into this custom set, rather than relying solely on structured ICD code mapping.

## Model Selection

Four prominent LLMs were selected for this study based on their reputation for processing unstructured data in health care contexts, along with BioBERT, a domain-specific biomedical model chosen for its specialization in biomedical text:

- GPT-3.5 (OpenAI): A general-purpose LLM trained on diverse datasets.
- GPT-4o (OpenAI): An advanced iteration, GPT-4o ("o" for omni) offers enhanced capabilities, including support for both text and image inputs. It is available through a paid subscription and is designed to process complex inputs with greater intelligence and versatility.
- Llama 3.2 (Meta): An open-source model, it is designed for efficient processing. The 70B parameter version was deployed locally using 4-bit quantization, available freely for research purposes.
- Gemini 1.5 (Google): Accessed through the Google Cloud Vertex AI platform. The model was configured with a temperature of 1 for creative outputs and set to generate responses of up to 8192 tokens to accommodate lengthy medical descriptions.
- BioBERT (dmis-lab): A specialized biomedical language model built on BERT architecture; it was accessed through the Hugging Face Transformers library. Unlike the LLMs, BioBERT was specifically pretrained on biomedical corpora for domain-specific tasks. The dmis-lab/biobert-base-cased-v1 model was deployed with a maximum token length of 128 and batch size of 8, and it was trained for 3 epochs with a weight decay of 0.01.

These models were chosen for their accessibility and integration capabilities in clinical data processing pipelines. Each model was tested on its ability to classify unstructured diagnosis descriptions into predefined categories.

### Prompt Design

Prompt design is crucial for enhancing the accuracy of LLM outputs, particularly in clinical applications. Working iteratively with the data science and clinical teams, we refined various prompt formulations, ultimately defining the following optimal prompt for diagnosis categorization:

> Given the following ICD-10 description or treatment note for a radiation therapy patient: {input}, select the most appropriate category from the predefined list: {Category list}. Respond only with the exact category name from the list. If the description indicates a benign tumor, categorize it as 'Benign.' Use 'Unknown' if it does not match any category. If the description overlaps multiple categories, select the closest matching category.

### Model Implementation

Each model was implemented in Python 3, but while most models were accessed via their respective cloud application programming interfaces, Meta's Llama 3.2 was implemented locally using Ollama. The models were integrated into the data processing pipeline to enable automated categorization of unstructured diagnosis data.

### Postprocessing and Category Standardization

To ensure consistency in model outputs, we implemented a 2-step category cleaning function in Python. The function first standardizes input by converting all text to lowercase and removing excess white space, then attempts to match the processed text against predefined categories. If no direct match is found, the function uses the difflib library to identify the matching category using string similarity metrics. This approach helps standardize variations in spellings and minor discrepancies in model outputs while maintaining classification accuracy. Cases without sufficient similarity to any predefined category are labeled as "No match."

### Diagnostic Category Verification

To evaluate model accuracy, the categorized diagnoses were independently reviewed by oncology experts (DLS and CLB). Each model's performance was quantified by measuring alignment between its outputs and expert classifications, with accuracy calculated based on agreement rates. Patterns of misclassification were examined to highlight common errors and areas for potential model improvement.

### Validation

We evaluated model performance using two primary metrics: accuracy and weighted macro $F_1$-score. Accuracy was calculated by directly comparing each model's predicted diagnosis to the true label, without adjusting for class frequency. Weighted macro $F_1$-score provided a complementary measure that balances precision and recall across all diagnosis categories while assigning greater influence to frequently occurring diagnoses via sample weights. This ensured that performance on common categories meaningfully impacted the overall score while still considering the model's ability to classify less frequent diagnoses.

To assess the statistical robustness of our estimates, we computed 95% confidence intervals for both metrics using nonparametric bootstrapping. This resampling approach quantifies the uncertainty around each point estimate, allowing for more reliable comparisons between models.

### Ethical Considerations

This study was approved by the institutional review board of the University of Tennessee Health Science Center (institutional review board number 16-04888-XP) as exempt under 45 CFR 46.104(d)(4), as it involved the secondary analysis of deidentified clinical data. Because the data were fully deidentified and the research involved no direct patient interaction, informed consent was not required. Data privacy was maintained through secure institutional servers and restricted access for authorized personnel only.

## Results

From our dataset of 3456 patients with cancer, we analyzed 762 unique diagnoses: 326 structured ICD code descriptions and 436 free-text entries from clinical notes. Each unique diagnosis was mapped to one category based on expert-defined criteria, and the frequency of each diagnosis across patient records was tallied. Table 1 summarizes the number of unique diagnoses per category alongside their total frequency in the dataset.

Model evaluation revealed varied performance patterns across data formats (Table 2). BioBERT achieved the highest weighted macro $F_1$-score on ICD codes (84.2, 95% CI 78.4-89.7), demonstrating the effectiveness of biomedical-specific pretraining for interpreting structured data. Its ICD code accuracy was also strong at 90.8 (95% CI 87.7-93.9). However, for free-text diagnoses, accuracy dropped to 81.6 (95% CI 78.0-85.3) and weighted macro $F_1$-score to 61.5 (95% CI 53.3-69.1). This drop reflects challenges in processing unstructured clinical narratives, even with strong performance on standardized medical terminology. BioBERT particularly struggled with cases involving abbreviations and multicondition diagnoses.

**Table 1.** Distribution of cancer diagnosis categories defined by oncology experts, showing the number of unique diagnosis descriptions and their total frequency across 3456 patient records.

| Cancer category | Total frequency, n | Unique diagnoses, n |
| --- | --- | --- |
| Breast | 539 | 23 |
| Lung or thoracic | 516 | 67 |
| Head and neck | 485 | 133 |
| Metastasis | 466 | 94 |
| Gastrointestinal | 329 | 94 |
| Prostate | 304 | 3 |
| Benign | 178 | 32 |
| Gynecologic | 174 | 81 |
| Central nervous system | 155 | 77 |
| Hematologic | 91 | 44 |
| Soft tissue | 59 | 24 |
| Genitourinary | 58 | 22 |
| Skin | 56 | 33 |
| Unknown | 46 | 35 |
| Total | 3456 | 762 |

**Table 2.** Accuracy and weighted macro $F_1$-score of different language models in categorizing medical diagnoses for ICD[a] code and free-text analysis (mean, 95% CI).

| Model | ICD code analysis | | Free text analysis | |
| --- | --- | --- | --- | --- |
| | Accuracy (%) | Weighted macro $F_1$-score (%) | Accuracy (%) | Weighted macro $F_1$-score (%) |
| GPT-3.5 | 88.9 (85.6-92.3) | 61.6 (55.9-79.7) | 75.9 (71.8-79.8) | 52.5 (46.5-58.4) |
| GPT-4o | 90.8 (87.7-93.9) | 81.2 (74.9-87.2) | 81.9 (78.0-85.1) | 71.8 (62.1-78.1) |
| Gemini | 81.0 (76.7-85.0) | 70.2 (62.5-76.5) | 64.2 (59.6-68.8) | 51.6 (43.9-57.8) |
| Llama | 77.6 (73.0-81.9) | 64.5 (54.6-73.7) | 64.2 (59.4-68.8) | 43.9 (38.0-50.3) |
| BioBERT | 90.8 (87.7-93.9) | 84.2 (78.4-89.7) | 81.6 (78.0-85.3) | 61.5 (53.3-69.1) |

[a]ICD: International Classification of Diseases.

Among the LLMs, GPT-4o matched BioBERT's ICD code accuracy at 90.8 (95% CI 87.7-93.9) and achieved the second-highest weighted macro $F_1$-score for ICD codes at 81.2 (95% CI 74.9-87.2). In free-text performance, GPT-4o outperformed BioBERT in weighted macro $F_1$-score, reaching 71.8 (95% CI 62.1-78.1) versus BioBERT's 61.5 (95% CI 53.3-69.1), while maintaining slightly higher accuracy at 81.9 (95% CI 78.0-85.1) compared to 81.6 (95% CI 78.0-85.3).

GPT-3.5 delivered competitive accuracy on ICD code descriptions (88.9, 95% CI 85.6-92.3) but showed a sharper decline for free-text diagnoses (75.9, 95% CI 71.8-79.8). Its weighted macro $F_1$-score dropped from 61.6 (95% CI 55.9-79.7) for ICD codes to 52.5 (95% CI 46.5-58.4) for free text, highlighting limitations in unstructured note processing. Gemini and Llama both trailed the top performers, with ICD code accuracies of 81.0 (95% CI 76.7-85.0) and 77.6 (95% CI 73.0-81.9), respectively, and further reductions for free text (64.2, 95% CI 59.6-68.8 for both). Weighted macro $F_1$-score followed the same pattern, with Gemini scoring 70.2 (95% CI 62.5-76.5) for ICD codes and 51.6 (95% CI 43.9-57.8) for free text, and Llama scoring 64.5 (95% CI 54.6-73.7) and 43.9 (95% CI 38.0-50.3), respectively.

Across all models, performance was consistently higher for structured ICD codes than for free-text diagnoses,

underscoring the inherent challenges of unstructured clinical narratives for language models in health care contexts.

Error analysis revealed distinct misclassification patterns across structured (325 ICD codes) and unstructured (435 free text) diagnoses. Table 3 compares the top two misclassification patterns for each model across both formats. For ICD codes, BioBERT primarily misclassified benign tumors as central nervous system (CNS) (6 cases), while other models struggled to differentiate between adjacent body regions and systems, such as Head and Neck cancers and Breast cancers. Specifically, GPT-3.5 misclassified benign tumors as CNS (13 cases) and Soft Tissue (3 cases). GPT-4o had similar struggles with benign tumors misclassified as CNS (10 cases) and Soft Tissue (2 cases).

Table 3. Top two misclassification patterns for each model across ICD[a] code and free-text analysis.

| Model | ICD code analysis | | | Free-text analysis | | |
|---|---|---|---|---|---|---|
| | True category | Misclassified as | Count | True category | Misclassified as | Count |
| GPT-3.5 | | | | | | |
| | Benign | CNS[b] | 13 | Metastasis | CNS | 14 |
| | Benign | Soft tissue | 3 | Metastasis | Unknown | 10 |
| GPT-4o | | | | | | |
| | Benign | CNS | 10 | Metastasis | Lung or thoracic | 14 |
| | Benign | Soft tissue | 2 | Unknown | Breast | 9 |
| Gemini | | | | | | |
| | Skin | Head and neck | 11 | Metastasis | Lung or thoracic | 25 |
| | Benign | CNS | 10 | Gastrointestinal | Genitourinary | 20 |
| Llama | | | | | | |
| | Breast | Gynecologic | 8 | Metastasis | CNS | 12 |
| | Hematologic | Head and neck | 6 | Unknown | Head and neck | 10 |
| BioBERT | | | | | | |
| | Benign | CNS | 6 | Unknown | Metastasis | 6 |
| | Head and neck | Skin | 2 | Unknown | Lung or thoracic | 5 |

[a]ICD: International Classification of Diseases.
[b]CNS: central nervous system.

In free-text analysis, all models struggled significantly with "Unknown" categorizations, with 6-12 cases misclassified variously as Metastasis, CNS, Lung or Thoracic, or Head and Neck. For instance, GPT-3.5 misclassified Unknown as CNS (14 cases) and Metastasis as Unknown (10 cases). Similarly, GPT-4o misclassified Unknown as Breast (9 cases) and Metastasis as Lung or Thoracic (14 cases). Metastatic disease posed particular difficulties in free text, frequently miscategorized as CNS or Lung or Thoracic (5-9 cases). For example, Gemini misclassified Metastasis as Lung or Thoracic (25 cases) and CNS (13 cases). This pattern suggests that the models face greater challenges in interpreting ambiguous or complex descriptions in free text compared to standardized ICD codes, as they often struggle to accurately identify Unknown or ambiguous diagnoses in the absence of a clear structure.

## Discussion

### Principal Findings

This study evaluated the performance of 4 LLMs and 1 specialized biomedical language model in categorizing cancer diagnoses from both structured ICD codes and free-text descriptions. The results revealed several valuable insights for clinical applications of language models.

BioBERT's superior performance on ICD codes (accuracy of 90.8%, weighted macro $F_1$-score of 84.2%) highlights the importance of domain-specific pretraining for structured medical data, consistent with recent studies in biomedical classification [12,13]. In free-text analysis, GPT-4o demonstrated performance comparable to BioBERT, achieving an accuracy of 81.8% and a weighted macro $F_1$-score of 82.2%, closely matching BioBERT's accuracy of 81.6% and weighted macro $F_1$-score of 84.2%. This suggests that while BioBERT holds a slight edge in weighted macro $F_1$-score, GPT-4o performs exceptionally well in free-text classification, showing that general-purpose LLMs like GPT-4o can effectively manage the complexity and variability of narrative clinical descriptions. The performance gap between the two models in free-text classification is relatively small, indicating that GPT-4o can perform nearly on par with BioBERT in unstructured clinical tasks. This is particularly noteworthy given that GPT-4o is a general-purpose model and was not specifically trained on biomedical corpora. Its strong performance indicates that modern general LLMs may be sufficiently robust for certain clinical NLP tasks, especially when domain-specific models are unavailable or when flexibility across different text types is needed.

Error analysis revealed significant challenges for language models in accurately categorizing benign tumors and metastasis. Benign tumors, while present in many diagnostic codes, often lack specific terminology indicating "benign," which makes it difficult for models to differentiate these cases. This limitation in the models' ability to accurately identify and classify benign diagnoses suggests the need for more specialized training that incorporates the nuances of medical coding and terminology, especially when explicit markers like "benign" are absent.

Metastasis presents an even greater challenge for accurate classification. In many cases, patients have multiple diagnosis codes that could be classified under both cancer types and metastasis, depending on the body regions affected. For instance, metastasis may involve various organs or systems that are not explicitly labeled as metastatic in the diagnosis code. In this study, the expert team chose to treat metastasis as a distinct category, regardless of the body locations involved. This decision increased the complexity for the language models, which needed to establish thresholds to accurately identify and categorize metastasis, even when the diagnosis lacked explicit mention of its metastatic nature. As a result, distinguishing between primary cancer diagnoses and metastasis proved to be a particularly demanding task for the models, emphasizing the need for improved context understanding and refined classification thresholds to better manage ambiguous or overlapping medical descriptions.

In free-text analysis, distinguishing between appropriate Unknown classifications and specific cancer categories emerged as a significant challenge. Models often attempted to force-classify vague or incomplete clinical descriptions into specific cancer categories, rather than correctly assign them as Unknown. For instance, cases marked as Unknown by clinical experts due to insufficient information (like "right" or "pelvis" without context) were often misclassified by models into specific cancer types based on partial pattern matching. This tendency to overclassify ambiguous cases suggests that additional preprocessing steps may be needed to effectively identify and handle low-quality or incomplete data entries before classification.

Unlike structured ICD codes, free-text entries often include abbreviations, shorthand, and nonstandardized phrasing (eg, "b/1 GBM," "left RMT"), which are difficult to interpret without domain-specific knowledge. Many notes also lack explicit diagnostic terms, instead relying on symptomatic or anatomical references that require inference to determine the underlying diagnosis. Fragmented grammar and co-reference issues (eg, determining which condition is primary) further obscure the clinical meaning. These factors contribute to reduced model accuracy and highlight the need for enhanced preprocessing or confidence-based filtering of ambiguous inputs. Although current performance may be sufficient for administrative or research tasks, expert validation remains essential for high-stakes clinical use, particularly when interpreting complex or ambiguous free-text entries.

Given the observed misclassifications in high-risk categories such as Metastasis and Unknown, relying solely on automated outputs could pose clinical risks. To mitigate this, future implementations should incorporate confidence thresholds to flag uncertain predictions, rule-based validation checks to catch implausible outputs, and human-in-the-loop systems to review ambiguous or low-confidence cases. Such hybrid approaches would balance automation efficiency with the need for clinical reliability, especially when dealing with unstructured or incomplete documentation.

These findings have meaningful implications for real-world adoption. The high accuracy on structured ICD codes indicates that models like BioBERT and GPT-4o can support EHR automation and administrative tasks such as billing and documentation with minimal manual input. However, the lower performance on free-text inputs underscores the need for caution in clinical decision support, where unstructured data are common and precision is critical. A hybrid approach—automating routine cases while flagging ambiguous ones for expert review—may offer the most practical balance between efficiency and safety.

Our study has several limitations that may impact the generalizability of the findings. Because the dataset was drawn from a single radiation oncology department within one institution, the models were trained and evaluated on documentation patterns specific to that clinical environment. In real-world applications, model performance may vary across health care systems that use different EHR platforms, coding practices, clinical documentation styles, or language contexts. Additionally, patient populations can differ demographically and clinically, which may influence diagnostic terminology and disease representation. The variation in sample size across cancer types, especially rare diagnoses, could also skew performance metrics. Furthermore, our analysis was limited to English-language records from a specific geographic region, potentially overlooking linguistic and cultural differences in medical documentation. Future work should validate these models using data from multiple institutions and more diverse patient populations to assess their robustness across settings.

LLMs can reflect biases in their training data, leading to uneven performance across demographic groups or rare diagnoses. They may also generate clinically plausible but incorrect outputs, posing risks if used without oversight. To ensure safe and equitable use in health care, these models should be validated on diverse populations and implemented with human oversight to prevent overreliance on automation.

## Conclusion

This study highlights both the promise and the challenges of using language models for automated cancer diagnosis classification. BioBERT and GPT-4o achieved strong performance, with BioBERT reaching 90.8% accuracy and a weighted macro $F_1$-score of 84.2% for ICD codes and GPT-4o achieving equivalent ICD accuracy (90.7%) with a slightly lower weighted macro $F_1$-score. In free-text classification, GPT-4o attained an accuracy of 81.8% and a weighted macro $F_1$-score of 82.2%, closely matching BioBERT's accuracy of 81.6% and weighted macro $F_1$-score of 84.2%. These results show considerable promise for clinical applications, particularly with structured data, while also demonstrating that general-purpose LLMs like GPT-4o can approach the performance of domain-specific models in unstructured tasks.

Nonetheless, significant challenges remain in free-text classification, especially for complex categories such as metastasis and ambiguous Unknown cases. These limitations underscore the need for enhanced preprocessing, confidence-based filtering, and human-in-the-loop validation to ensure safe clinical adoption. With further fine-tuning of domain-specific models like BioBERT, improvements in ICD coding systems, and hybrid approaches that combine automation with expert oversight, more robust and reliable tools can be developed for real-time prediction in health care workflows. Future work should prioritize improving free-text classification, refining thresholds for complex diagnoses, and validating across diverse, multiinstitutional datasets to ensure generalizability and clinical robustness.

### Acknowledgments
We thank the University of Tennessee IT team for their support with data access and infrastructure. This research was supported by a grant from the Tennessee Department of Health. Generative artificial intelligence tools (such as ChatGPT) were used to assist in language editing and formatting of the manuscript. All content was reviewed and verified by the authors for accuracy and integrity.

### Data Availability
The data that support the findings of this study are available upon reasonable request from the corresponding author. The dataset is not publicly available due to institutional privacy policies. The code used for data preprocessing, model evaluation, and statistical analysis in this study is publicly available at [14].

### Authors' Contributions
SH: conceptualization, methodology, data analysis, writing–original draft.
RR: data analysis, validation.
CLB: conceptualization, clinical validation, writing–review & editing.
LC: data analysis, validation.
FAK: data preparation, writing–review & editing.
JAZ: data preparation, writing–review & editing.
RLD: conceptualization, clinical validation.
DLS: conceptualization, clinical validation, review, editing, supervision, and funding acquisition.
ASN: conceptualization, writing–review & editing, supervision, and funding acquisition.

### Conflicts of Interest
None declared.

### References
1. Jensen PB, Jensen LJ, Brunak S. Mining electronic health records: towards better research applications and clinical care. Nat Rev Genet. May 2, 2012;13(6):395-405. [doi: 10.1038/nrg3208] [Medline: 22549152]
2. Yuan Q, Cai T, Hong C, et al. Performance of a machine learning algorithm using electronic health record data to identify and estimate survival in a longitudinal cohort of patients with lung cancer. JAMA Netw Open. Jul 1, 2021;4(7):e2114723. [doi: 10.1001/jamanetworkopen.2021.14723] [Medline: 34232304]
3. Panahiazar M, Taslimitehrani V, Pereira N, Pathak J. Using EHRs and machine learning for heart failure survival analysis. Stud Health Technol Inform. 2015;216:40-44. [doi: 10.3233/978-1-61499-564-7-40] [Medline: 26262006]
4. Ford E, Carroll JA, Smith HE, Scott D, Cassell JA. Extracting information from the text of electronic medical records to improve case detection: a systematic review. J Am Med Inform Assoc. Sep 2016;23(5):1007-1015. [doi: 10.1093/jamia/ocv180] [Medline: 26911811]
5. Shaban-Nejad A, Michalowski M, Bianco S. Creative and generative artificial intelligence for personalized medicine and healthcare: hype, reality, or hyperreality? Exp Biol Med (Maywood). Dec 2023;248(24):2497-2499. [doi: 10.1177/15353702241226801] [Medline: 38311873]
6. Yang H, Xiang K, Ge M, Li H, Lu R, Yu S. A comprehensive overview of backdoor attacks in large language models within communication networks. IEEE Netw. 2024;38(6):211-218. [doi: 10.1109/MNET.2024.3367788]
7. Park YJ, Pillai A, Deng J, et al. Assessing the research landscape and clinical utility of large language models: a scoping review. BMC Med Inform Decis Mak. Mar 12, 2024;24(1):72. [doi: 10.1186/s12911-024-02459-6] [Medline: 38475802]
8. Abu-Ashour W, Emil S, Poenaru D. Using artificial intelligence to label free-text operative and ultrasound reports for grading pediatric appendicitis. J Pediatr Surg. May 2024;59(5):783-790. [doi: 10.1016/j.jpedsurg.2024.01.033] [Medline: 38383177]

9. Kaufman DR, Sheehan B, Stetson P, et al. Natural language processing-enabled and conventional data capture methods for input to electronic health records: a comparative usability study. JMIR Med Inform. Oct 28, 2016;4(4):e35. [doi: 10.2196/medinform.5544] [Medline: 27793791]
10. Houssein EH, Mohamed RE, Ali AA. Machine learning techniques for biomedical natural language processing: a comprehensive review. IEEE Access. 2021;9:140628-140653. [doi: 10.1109/ACCESS.2021.3119621]
11. Hsu E, Malagaris I, Kuo YF, Sultana R, Roberts K. Deep learning-based NLP data pipeline for EHR-scanned document information extraction. JAMIA Open. Jul 2022;5(2):ooac045. [doi: 10.1093/jamiaopen/ooac045] [Medline: 35702624]
12. Chen S, Li Y, Lu S, et al. Evaluating the ChatGPT family of models for biomedical reasoning and classification. J Am Med Inform Assoc. Apr 3, 2024;31(4):940-948. [doi: 10.1093/jamia/ocad256] [Medline: 38261400]
13. Tian S, Jin Q, Yeganova L, et al. Opportunities and challenges for ChatGPT and large language models in biomedicine and health. Brief Bioinformatics. Nov 22, 2023;25(1). [doi: 10.1093/bib/bbad493]
14. Hashtarkhani/cancer_classification. GitHub. URL: https://github.com/hashtarkhani/Cancer_Classification [Accessed 2025-09-29]

## Abbreviations

**AI:** artificial intelligence
**BERT:** bidirectional encoder representations from transformers
**CNS:** central nervous system
**EHR:** electronic health record
**ICD:** International Classification of Diseases
**LLM:** large language model
**NLP:** natural language processing